\documentclass{article}
\usepackage{ijcai26}

\usepackage{times}
\usepackage{soul}
\usepackage{url}
\usepackage[hidelinks]{hyperref}
\usepackage[utf8]{inputenc}
\usepackage[small]{caption}
\usepackage{graphicx}
\usepackage{amsmath}
\usepackage{amsthm}
\usepackage{booktabs}
\usepackage{algorithm}
\usepackage{algorithmic}
\usepackage[switch]{lineno}
\usepackage{todonotes}
\usepackage{amssymb}
\usepackage{multirow}
\usepackage{pifont}
\usepackage[table]{xcolor}
\usepackage{array}
\usepackage{tikz}
\usepackage{paralist}

\definecolor{afgray}{gray}{0.92}

\title{
Looking Beyond Accuracy: A Holistic Benchmark of ECG Foundation Models
}

\author{
Francesca Filice$^1$
\and
Edoardo De Rose$^1$\and
Simone Bartucci$^{1,2,3}$\and
Francesco Calimeri$^{1,3}$\And
Simona Perri$^{1}$\\
\affiliations
$^1$Department of Mathematics and Computer Science, University of Calabria, Italy\\
$^2$Department of Computer, Control and Management Engineering “Antonio Ruberti”, Sapienza University of Rome, Italy\\
$^3$DLVSystem Srl, Rende, Italy\\
\emails
\{francesca.filice, edoardo.derose, simone.bartucci, francesco.calimeri, simona.perri\}@unical.it}

\begin{document}

\maketitle

\begin{abstract}
The electrocardiogram (ECG) is a cost-effective, highly accessible and widely employed diagnostic tool. 
With the advent of Foundation Models (FMs), the field of AI-assisted ECG interpretation has begun to evolve, as they enable model reuse across different tasks by relying on \textit{embeddings}. 
However, to responsibly employ FMs, it is crucial to rigorously assess to which extent the embeddings they produce are generalizable, particularly in error-sensitive domains such as healthcare.
Although prior works have already addressed the problem of benchmarking ECG-expert FMs, they focus predominantly on the evaluation of downstream performance.
To fill this gap, this study aims to find an in-depth, comprehensive benchmarking framework for FMs, with a specific focus on ECG-expert ones.
To this aim, we introduce a benchmark methodology that complements performance-based evaluation with representation-level analysis, leveraging SHAP and UMAP techniques.
Furthermore, we rely on the methodology for carrying out an extensive evaluation of several ECG-expert FMs pretrained via state-of-the-art techniques over different cross-continental datasets and data availability settings; this includes ones featuring data scarcity, a fairly common situation in real-world medical scenarios.
Experimental results show that our benchmarking protocol provides a rich insight of ECG-expert FMs' embedded patterns, enabling a deeper understanding of their representational structure and generalizability.
\end{abstract}

\section{Introduction}
Artificial Intelligence (AI) is rapidly evolving in recent years.
The advent of Foundation Models (FMs) represented a key milestone in this domain: in fact, by learning compact representations (known as \textit{embeddings}) from complex input data, FMs offer several advantages.
These include, for instance, reduced storage and training times requirements; indeed, training a downstream model over an embedded dataset rather than a high-dimensional one demands significantly lower resources.
This is precisely how FMs have unlocked model reusability across different tasks or domains, thereby avoiding the development of an \textit{ad-hoc} architecture.
FMs exhibit substantial versatility, a property that particularly benefits the medical domain, where their adoption is promptly expanding~\cite{wong:leveraging_FMs_and_LLMs_in_medical_AI}.
Example tasks on which FMs excel include medical imaging segmentation~\cite{ma:medsam}, classification of clinical conditions or pathologies~\cite{moor:FMs_for_generalist_medical_AI}, multimodal question answering and report generation~\cite{sellergren:medgemma}.
Interestingly, FMs application is growing even over ECG timeseries analysis~\cite{mckeen:ecg-fm,coppola:hubert-ecg,kuba:ecg-jepa,li:ecgfounder}, also due to the accessibility and cost-effectiveness of ECG as a diagnostic tool, which makes it particularly well suited for AI-based approaches.
Nonetheless, employing FMs in such an error-sensitive domain as ECG interpretation demands a thorough awareness of their generalization capabilities and, in particular, their limitations. 
In fact, if a FM fails to generalize properly when embedding the provided input data, its downstream performance may be unreliable, potentially leading to urgent consequences such as \textit{underdiagnosis}~\cite{yang:underdiagnosis_1,bahre:underdiagnosis_2}. 
This highlights that the generalization capabilities of FMs must be accurately assessed before their actual deployment~\cite{jin:fairmedfm}.
However, performing such a comprehensive evaluation is rather challenging, as it requires to identify whether the FM's pattern-recognition is clinical-oriented (i.e., whether it filters clinically meaningful patterns) rather than based on exploiting spurious dataset-related correlations, which may compromise generalization especially for overparametrized models~\cite{ye:spurious_correlations_survey}.
To investigate this, performance analysis alone is insufficient, as it does not adequately characterize to which extent the generated embeddings are: $(1)$ general-purpose; $(2)$ informative (i.e., semantically rich); $(3)$ data-shift invariant.

This work proposes a modular and systematic methodology to perform a holistic (i.e., both performance-wise and representation-wise) benchmarking of FMs for ECG data.
The FMs are employed as frozen embedding extractors, thus resulting in a zero-shot setting, which assures that the embedded representation are evaluated as they are right after the pretraining stage.
The proposed approach is cross-strategy, cross-continental and multi-scale: in fact, the encompassed FMs differ in both up-to-date architecture and pretraining strategy and are evaluated across balanced size-varying subsets of geographically diverse ECG signal datasets.
As such, this benchmarking delivers a comprehensive and State-Of-The-Art–aligned assessment of ECG-expert FMs.
To the best of our knowledge, this is the first approach that exhaustively benchmarks ECG-expert FMs across all the dimensions discussed before.
Thus, to sum up, our contribution is threefold:
\begin{compactitem}
\item 
    We develop an ECG-expert FM benchmarking methodology that complements standard performance evaluation with a representation-wise one, closing the gap in ECG-expert FM representation benchmarking;
\item 
    We provide an empirically validated baseline for ECG-expert FM benchmarking and a ready-to-use framework for the research community.
\item 
    Upon publication, we will publicly release the full codebase to enable the research community to implement, reproduce, and extend our benchmarking framework.
\end{compactitem}
With this work, we aim to develop a comprehensive understanding of ECG-expert FMs’ generalization, raise awareness of their limitations, and promote more accessible benchmarking baselines through open-source development.

\section{Related Work} \label{sec:related_works}
Previous studies have benchmarked FMs on ECG data. 
For instance, some studies evaluate ECG-expert FMs across multiple datasets and downstream tasks and compare their performance against supervised baseline models~\cite{almasud:benchamrking_ecg_fms_a_reality,lunelli:benchecg}. 
Others introduce the incidence of dataset scale by analyzing model performance under varying amounts of training data in a cross-dataset setting, where, in turns, one dataset is left out for testing purposes, while the remaining ones are employed for training~\cite{wan:openecg}. 
Finally, previous research also broadens the analysis to language models and general-purpose FMs, while limiting the employment of a single ECG-expert FM~\cite{xu:an_ECG_multitask_benchmark}.

Nonetheless, none of these works include a representation-wise evaluation in their benchmark: instead, they focus on performances over downstream tasks.
However, recent non-ECG-centered works argue that benchmarking FMs should go beyond downstream task performance, thus explicitly evaluating both the structure and alignment of learned representations.
For example, FIND~\cite{zou:find} proposes a benchmark for FMs that explicitly targets the quality of their learned embedding space, introducing interleaved retrieval and grounding tasks to assess alignment across modalities and granularities over visual and language FMs. 
FairMedFM~\cite{jin:fairmedfm} further demonstrates the importance of evaluating FMs at the representation level, introducing metrics that assess subgroup separability and representation fairness in embedding spaces.

\section{Methods}
Motivated by the representation-aware FM benchmarking philosophy that is still missing in State-Of-The-Art ECG-expert FM benchmarks (as discussed in Section~\ref{sec:related_works}), in this work we extend the notion of such representation-aware benchmarking to the ECG domain.
Specifically, we propose a benchmarking protocol for ECG-expert FMs that incorporates representation-wise evaluation through SHAP and UMAP-based evaluation~\cite{lundberg:shap,mcinnes:umap}, fulfilling the gap in the State-Of-The-Art proposals.
By doing this, we are able to examine interpretability, feature attribution and inner structure alongside standard performance metrics, thus providing a full view of the embedding properties.
The benchmark comprehends 4 different geographically spread datasets and 4 different FMs architectures.
For each benchmarked FM, the pipeline unfolds as follows:
\begin{compactenum}
    \item {\it Embedding extraction}: We employ the considered FM as a  {\it frozen embedding extractor} (i.e., it is not finetuned at all), thus obtaining the embedded representation of the preprocessed input data. 
    By doing that, we make the embeddings' quality dependent exclusively from the FM's pretraining strategy.
    This concurs into assuring fairness during comparison.
    \item {\it Linear probing}: with the so obtained embeddings, we train  5 different lightweight classifiers (Cs) (i.e., XGBoost, Decision Tree, Random Forest, Logistic Regression, Multi Layer Perceptron).
    \item {\it Performance evaluation}: we analyze classification performances via F1 score over a 15-fold cross-validation scheme.
    This evaluation assesses whether the embeddings provide sufficient information to effectively support classifiers' training for accurate performance.
    \item {\it Representation evaluation}: we analyze the informativeness of the embedded representations and the geometry of the embedding space through SHAP feature ranking and embedding space visualization and evaluation via UMAP and cluster-based metrics. 
\end{compactenum}
The benchmarking procedure is carried over 4 different dataset scaled from less than 500 to over 5000 samples.
This stress-tests FMs generalization capabilities even under data-scarce conditions (a realistic scenario, especially in privacy-restricted domains such as medicine), thus making the present study timely and relevant for real-world settings and strengthening its overall scope.

\begin{table}[t]
    \centering
    \begin{tabular}{llrr}
        \toprule
        \textbf{Dataset}  & \textbf{Origin} & \textbf{\# classes} & \textbf{\# samples}  \\
        \midrule
        PTX               & Europe         &  71    & 21.837     \\
        C15            & America        &  7     & 345.779    \\
        GEO              & America        &  67    & 10.344     \\
        CHN       & Asia           &  30    & 10.247     \\
        \bottomrule
    \end{tabular}
    \caption{\vspace{-1ex}Datasets details.}
    \label{tab:datasets_details}
\end{table}

\subsection{Datasets}
The present benchmarking is conducted over the following datasets containing 12-leads ECG signals: Georgia (GEO)~\cite{reyna:physionet2021}, CODE-15\%  (C15)~\cite{ribeiro:code15}, PTB-XL (PTX)~\cite{wagner:ptbxl,reyna:physionet2021}, Chapman-Shaoxing+Ningbo (CHN)~\cite{zheng:chapman,zheng:ningbo,reyna:physionet2021}.
These datasets vary by geographical origin, number of samples and number of classes, as listed in Table~\ref{tab:datasets_details}.

\subsection{Foundation Models}
We benchmark 4 different ECG-expert FMs, each varying in architectural design and pretraining methodology.
This comparison aims to comprehensively cover up-to-date FM architectures and pretraining techniques.
The encompassed FMs are listed below.

\paragraph{ECG-FM.} 
ECG-FM~\cite{mckeen:ecg-fm} is a self-supervised architecture composed of a multi-layer CNN feature extractor concatenated to a BERT-like transformer encoder~\cite{choi:ecgbert}.
Its pretraining pipeline is based on both contrastive learning and masking strategies.
Specifically, contrastive learning was implemented by treating temporally adjacent ECG segments as positive pairs and non-adjacent segments as negative pairs.
In parallel, the masking strategy consists in masking contiguous spans of latent CNN representations, thereby inducing an inter-layer masking process.

\paragraph{ECGFounder.}
ECGFounder~\cite{li:ecgfounder} distinguishes itself through two key characteristics. 
First, it is built upon the RegNet architecture~\cite{radosavovic:regnet}, which is designed to predict optimal network widths and depths, thereby ensuring computational efficiency and adaptability. 
Second, its pretraining pipeline adopts a less conventional multi-label classification strategy rather than a single-label one: this strongly reflects real-world clinical scenarios in which ECG recordings are frequently associated with multiple diagnostic labels.

\paragraph{HuBERT-ECG.}
As its name suggests, HuBERT-ECG~\cite{coppola:hubert-ecg} is derived from the BERT~\cite{choi:ecgbert} architecture. 
It consists of a convolutional feature extractor followed by a transformer encoder. 
A distinctive aspect of HuBERT-ECG’s pretraining pipeline is the use of self-supervised label induction via k-means clustering, which is subsequently followed by a masking approach. 
Moreover, through transfer learning from the base model (HuBERT-base), the authors derive two additional FMs of different scales (HuBERT-small and HuBERT-large).
In this work, we benchmark each of these differently sized versions.

\paragraph{ECG-JEPA.}
The most noteworthy characteristic of ECG-JEPA~\cite{kuba:ecg-jepa} lies in its foundations on the Joint-Embedding Predictive Architecture (JEPA) framework~\cite{assran:jepa}, which consists in predicting masked information directly in the embedding space (that is, from an embedded representation to another) rather than operating on raw data, which are inherently more complex.
This strategy aims at training the FM to properly recognize and consequently ignore noisy low-level details.
Despite JEPA was originally developed for images, ECG-JEPA's authors adapt it to signals by modifying the Visual Transformers (ViT) architecture and implementing a signal-based masking strategy for ECG waveforms~\cite{kuba:ecg-jepa}.
While there exist different sized versions of ECG-JEPA (from smallest to largest: XS, S, B), in this work, we focus exclusively on benchmarking ViT-S following the authors' suggestion.

\subsection{Evaluating Performances}
By evaluating the classifiers' performance, we aim to assess whether the FMs produce embedded representations that are enough informative to support effective downstream training and, consequently, achieve satisfactory performance.
Specifically, to measure the classification performances we employed the F1 score; 
this metric is particularly relevant as it balances precision and recall: this is critical especially in the medical domain, where mispredictions may entail severe clinical risks~\cite{nafjan:AI_in_predictive_healthcare}, particularly w.r.t. false negatives (i.e., underdiagnosis).

\subsection{Evaluating Embedded Representations}
The embedding evaluation aims at investigating how the encompassed FMs ``understand'' clinical data, how they organize spatially their embeddings, which features they select/discard during the compression process, and to which extent their embeddings are generalizable.
To this aim, we study each FM's embedding space over the most influent features extracted from the best-performance-yielding dataset.
Algorithm~\ref{alg:shap_umap} outlines this methodology.
Specifically, for each FM, we  select the classifier which performs best over the encompassed datasets.
Subsequently, we rely on the selected classifier to extract the $50$ most influent features for each dataset (see Algorithm~\ref{alg:shap_umap}, line~\ref{top50}), thereby adopting a cross-dataset strategy.
We dimensionally reduce and plot the resulting embedding spaces through the UMAP technique.
Finally, we evaluate the embedded representations geometry considering both dataset-wise and label-wise perspectives.

\begin{algorithm}[t]
    \caption{Cross-dataset representation analysis}
    \label{alg:shap_umap}
    \textbf{Input}: $\mathcal{F}$ = set of $n$ FMs, $\mathcal{D}$ = set of $m$ datasets, $\mathcal{C}$ = set of $k$ classifiers \\
    \textbf{Output}: $\mathcal{U}$ = set of UMAP representations over the highest-performance-yielding dataset for each FM
    \begin{algorithmic}[1] 
        \STATE Let $\mathcal{U} = \varnothing$
        \FOR {$i \in \{1, \dots, n\}$}      
            \FOR {$j \in \{1, \dots, m\}$}  
                \STATE Let $C^*_{i,j} = \arg\max \mathrm{F1}(C_{i,j,r}) \; \forall r \in \{1, \dots, k\}$
            \ENDFOR
            \STATE Let $C^*_{i} =  \arg\max \mathrm{F1}(C^*_{i,j}) \; \forall j \in \{1, \dots, m\}$ \label{best_dataset_classifier}
            \STATE Let $D^{UMAP}_i =  \varnothing $
            \FOR {$j \in \{1, \dots, m\}$}  
                \STATE Let $D_j \in \mathcal{D}$, $FM_i \in \mathcal{F}$
                \STATE Let $D_{i,j} = FM_i(D_j)$  
                \STATE Let $\phi_{i,j} = shap(C^*_i, D_{i,j})$
                \STATE Let $\phi^{50}_{i,j} = top50features(\phi_{i,j})$ \label{top50}
                \STATE $D^{UMAP}_i \leftarrow \phi^{50}_{i,j}(D_j)$
            \ENDFOR
            \STATE $\mathcal{U} \leftarrow umap(D^{UMAP}_i)$
        \ENDFOR
        \STATE \textbf{return} $\mathcal{U}$
    \end{algorithmic}
\end{algorithm}

\paragraph{Ranking Feature Importance.} \label{par:shap_feature_ranking}
To rank the embeddings' feature importance for classification, we employed the SHAP technique.
This feature-ranking approach was selected because of its reliability and theoretical robustness, as it is both model-agnostic and grounded in mathematical theory: 
this makes it particularly suitable for our study as these properties guarantee a fair and stable comparison among different embeddings.
It is worth noting that, as shown in Algorithm~\ref{alg:shap_umap}, after selecting classifier $C^*_i$ based on the highest-performance-yielding dataset, we employ it to extract the top $50$ features via SHAP on every encompassed dataset.
This consists in a cross-dataset procedure which aims to examine \textit{whether} and, possibly, \textit{to which extent} the most influential features vary between the training dataset and other unseen datasets.
As already known to literature, models capable of extracting features invariant to domain-specific characteristics are more likely to capture the underlying semantic structure of the data and generalize to unseen domains~\cite{yoon:domain_generalization_2,khoee:domain_generalization_1}.
Motivated by this perspective, we propose to quantify the consistency of influential features across datasets as a practical and model-agnostic proxy for representation invariance, thereby providing an interpretable criterion to assess the generalization potential of FMs.
Thus, observing a greater overlap of influential features across the considered datasets suggests stronger generalization capabilities of $FM_i$.

\paragraph{Visualizing and Evaluating the Embedding Space.}
In order to represent the embedding space, we rely on the UMAP technique, as this constitutes a wide employed tool up to the State-Of-The-Art~\cite{torabizadeh:umap_cluster_1,keraghel:umap_cluster_2}.
Given that this methodology is known to be hyperparameter-sensitive~\cite{huang:umap_hyperparameters_sensitivity}, we tested over multiple hyperparameter combinations for each UMAP, in order to ensure that results remained consistent across different fixed configurations.
It is worth noting that, in this work, UMAP is not intended as a model selection criterion; instead, it is employed as a complementary, \textit{qualitative} tool to support representation-level analysis. 
For this reason, we augment the UMAP-based embedding space inspection with a set of \textit{quantitative} metrics, aiming at assessing the geometry of the embedding space prior to UMAP-based dimensionality reduction.
In particular, for an intra-cluster evaluation, we employ the \textit{k-Nearest Neighbour (kNN)}~\cite{cover:knn} metric, which quantifies the semantic similarities between a sample and its neighbours.
Conversely, for inter-cluster evaluation we compute two complementary metrics: \textit{centroid separation}~\cite{fisher:centroid}, which assesses whether classes are globally distinct in the embedding space by quantifying the normalized distance between centroids, and \textit{Adjusted Rand Index (ARI)}~\cite{hubert:ari}, which measures cluster separability (i.e., whether and, possibly, to which extent different clusters exhibit overlaps).
For the latter, we obtain the needed clusters via \textit{Gaussian Mixture Models (GMM)}~\cite{reynolds:gmm}. 
These metrics-based evaluation intends to quantitatively assess the label-level separability and dataset-level separability of the encompassed FMs' embedding spaces.
Label-level separability measures the ability of the embedding space to discriminate between clinical classes, while dataset-level separability captures domain-specific structure induced by differences in acquisition protocols, populations, preprocessing pipelines or pretraining techniques.
We thus provide a complementary two-perspective evaluation of the embeddings’ semantic structure, which is crucial for a thorough understanding of their informativeness.

\begin{table}[t]
    \centering
    \resizebox{\columnwidth}{!}{
    \begin{tabular}{l p{5cm} c}
        \toprule
        \textbf{Dataset} & \textbf{CD subclasses} & \textbf{AF presence} \\
        \midrule
        PTX & 1DAVB, CRBBB, IRBBB, LAnFB, CLBBB, ILBBB, IVCD, WPW, 2AVB, 3AVB &  \ding{55} \\
        C15\% & 1DAVB, RBBB, LBBB & $\checkmark$ \\
        GEO & 1DAVB, CRBBB, IRBBB, LAnFB & $\checkmark$ \\
        CHN & 1DAVB, CRBBB, IRBBB, LAnFB & $\checkmark$ \\
        \bottomrule
    \end{tabular}
    }
    \caption{Subclasses for the CD superclass and AF presence/absence for each dataset.
    Classes abbreviations unfold as follows: 1DAVB (first degree AV block), 2DAVB (second degree AV block), 3DAVB (third degree AV block), CLBBB (complete left bundle branch block), CRBBB (complete right bundle branch block), ILBBB (incomplete left bundle branch block), IRBBB (incomplete right bundle branch block), IVCD (non-specific intraventricular conduction disturbance (block)), LAnFB (left anterior fascicular block), LBBB (left bundle branch block), RBBB (right bundle branch block), WPW (Wolf-Parkinson-White syndrome).}
    \label{tab:cd_subclasses}
\end{table}

\begin{table*}[t]
    \centering
    \resizebox{\textwidth}{!}{

    \begin{tabular}{
    ll
    c>{\columncolor{afgray}}c
    c>{\columncolor{afgray}}c
    c>{\columncolor{afgray}}c
    c>{\columncolor{afgray}}c
    c>{\columncolor{afgray}}c
    c>{\columncolor{afgray}}c
    }
    \toprule
    \textbf{Dataset} & \textbf{Size} 
    & \multicolumn{2}{c}{\textbf{ECG-FM}} 
    & \multicolumn{2}{c}{\textbf{ECGFounder}} 
    & \multicolumn{2}{c}{\textbf{HuBERT-ECG small}} 
    & \multicolumn{2}{c}{\textbf{HuBERT-ECG base}} 
    & \multicolumn{2}{c}{\textbf{HuBERT-ECG large}} 
    & \multicolumn{2}{c}{\textbf{ECG-JEPA}} \\
\cmidrule(lr){3-14}
    & & \textbf{CD} & \textbf{AF} & \textbf{CD} & \textbf{AF} & \textbf{CD} & \textbf{AF} & \textbf{CD} & \textbf{AF} & \textbf{CD} & \textbf{AF} & \textbf{CD} & \textbf{AF} \\

    & 
    & \scriptsize IQR=0.03 & \scriptsize IQR=0.02
    & \scriptsize IQR=0.03 & \scriptsize IQR=0.01
    & \scriptsize IQR=0.04 & \scriptsize IQR=0.04
    & \scriptsize IQR=0.04 & \scriptsize IQR=0.03
    & \scriptsize IQR=0.02 & \scriptsize IQR=0.02
    & \scriptsize IQR=0.04 & \scriptsize IQR=0.04 \\

    \midrule

    \multirow{4}{*}{PTX}
     & XS  
     & 0.75  & -- 
     & \textbf{0.81}  & -- 
     & 0.75  & -- 
     & 0.73  & -- 
     & 0.74  & -- 
     & 0.76  & -- \\
     & S  
     & 0.77  & -- 
     & \textbf{0.83}  & -- 
     & 0.75  & -- 
     & 0.73  & -- 
     & 0.74  & -- 
     & 0.76  & -- \\
     & M  
     & 0.81  & -- 
     & \textbf{0.86}  & -- 
     & 0.77  & -- 
     & 0.76  & -- 
     & 0.76  & -- 
     & 0.78  & -- \\
     & L  
     & 0.83  & -- 
     & \textbf{0.87}  & -- 
     & 0.79  & -- 
     & 0.77  & -- 
     & 0.78  & -- 
     & 0.80  & -- \\

    \midrule
    
    \multirow{4}{*}{C15}
    & XS  
    & 0.91  & 0.94 
    & \textbf{0.93}  & \textbf{0.97} 
    & 0.83  & 0.84 
    & 0.79  & 0.77 
    & 0.84  & 0.82 
    & 0.82  & 0.67 \\
     & S  
    & 0.93  & 0.95 
    & \textbf{0.95}  & \textbf{0.97 }
    & 0.85  & 0.87
    & 0.84  & 0.82 
    & 0.84  & 0.84 
    & 0.81  & 0.74 \\
     & M  
    & \textbf{0.95}  & 0.95 
    & \textbf{0.95}  & \textbf{0.96} 
    & 0.85  & 0.88
    & 0.84  & 0.84 
    & 0.85  & 0.87 
    & 0.82  & 0.75 \\
     & L  
    & \textbf{0.95}  & 0.96 
    & \textbf{0.95}  & \textbf{0.96} 
    & 0.85  & 0.89
    & 0.83  & 0.86 
    & 0.84  & 0.88 
    & 0.83  & 0.77 \\
    
    \midrule
    
    \multirow{4}{*}{GEO}
     & XS  
     & 0.78  & 0.86 
     & \textbf{0.83}  & \textbf{0.88}
     & 0.65  & 0.73 
     & 0.65  & 0.74 
     & 0.63  & 0.75 
     & 0.65  & 0.63 \\
     & S  
     & 0.79  & 0.89 
     & \textbf{0.85}  & \textbf{0.89} 
     & 0.65  & 0.79 
     & 0.66  & 0.79 
     & 0.63  & 0.78 
     & 0.65  & 0.63 \\
     & M  
     & 0.83  & -- 
     & \textbf{0.88}  & -- 
     & 0.69  & -- 
     & 0.69  & -- 
     & 0.67  & -- 
     & 0.68  & -- \\
     & L  
     & -- & -- 
     & -- & -- 
     & -- & -- 
     & -- & -- 
     & -- & -- 
     & -- & -- \\
    
    \midrule
    
    \multirow{4}{*}{CHN}
     & XS  
     & 0.89  & 0.91 
     & \textbf{0.90}  & \textbf{0.93} 
     & 0.76  & 0.82 
     & 0.74  & 0.79 
     & 0.74  & 0.78 
     & 0.76  & 0.69 \\
     & S  
     & 0.89  & 0.92 
     & \textbf{0.91}  & \textbf{0.93} 
     & 0.75  & 0.83 
     & 0.74  & 0.82 
     & 0.75  & 0.81 
     & 0.76  & 0.73 \\
     & M  
     & 0.90  & 0.92 
     & \textbf{0.92}  & \textbf{0.93} 
     & 0.75  & 0.84 
     & 0.73  & 0.82 
     & 0.74  & 0.83 
     & 0.76  & 0.75 \\
     & L  
     & -- & -- 
     & -- & -- 
     & -- & -- 
     & -- & -- 
     & -- & -- 
     & -- & -- \\
     
    \bottomrule
    \end{tabular}
    }
    
    \caption{Classification performance over the CD and AF classes. 
    For each FM-Dataset-Size configuration, we report the performance of the best-performing classifier, expressed as the median F1 score across the 15 cross-validation folds.   
    In bold we mark the maximum median F1 score for each Dataset-Size-Class configuration. 
    At equal median F1 score, we select the one minimizing IQR.
    }
    \label{tab:cd_results}
\end{table*}

\begin{figure}[b]
    \centering
    \includegraphics[width=0.47\textwidth]{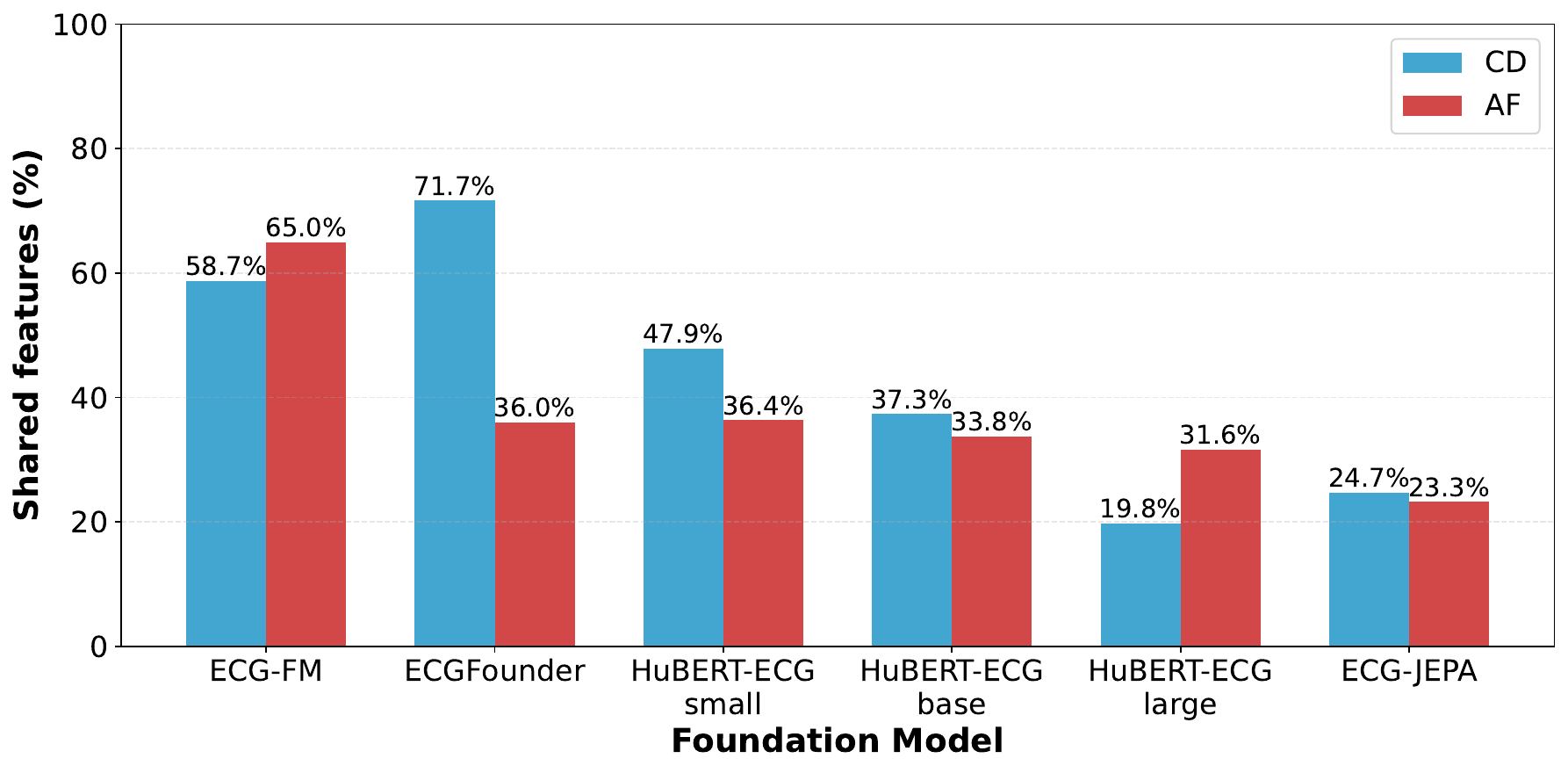}
    \caption{Shared top-50 features rates across datasets.
    Features are ranked via $C^*_i$ as discussed in Section~\ref{par:shap_feature_ranking} and described in Algorithm~\ref{alg:shap_umap}.
    }
    \label{fig:shap_features_percentage}
\end{figure}

\section{Experimental Settings}
Before being provided as input to any FM, each dataset is properly preprocessed according to the considered FM's preprocessing pipeline.
Moreover, given that the FMs' vector embeddings have shape $[p, q]$, where $p$ is the number of tokens and $q$ is the number of features, we pool such vectors across the first dimension, so to obtain a 1D vector of shape $[q]$.
We achieve this via two different pooling strategies: \textit{(1) Last-Token pooling (LST)}: we select the \textit{last} token of each feature; \textit{(2) Max Pooling}: we select the \textit{max} token for each feature.
During the linear probing phase, we perform a class-balanced 15-fold cross-validation in order to make the experiment statistically stable.
Moreover, we conduct an extensive hyperparameter search for each classifier via a grid search–based approach, so that optimal hyperparameter selection is ensured for downstream performance purposes.
The 4 different dataset ranges employed during experiments are the following: up to $499$ samples (XS), between $500$ and $2499$ samples (S), between $2500$ and $4999$ samples (M), over $5000$ samples (L).
For the L size, only two dataset (C15 and PTX) could be employed, as the remaining ones were not populated enough.
As for classification, we selected two labels: {\em CD} (Conduction Distrubance) and {\em AF} (Atrial Fribrillation).
CD is a composite class that groups different specific sub-classes, all relative to the Conduction Disturbance event.
Although the sub-classes composing the CD event vary slightly among datasets (Table~\ref{tab:cd_subclasses}), we consider this variability acceptable, as they consistently represent CD-related conditions, thereby broadening the scope of the evaluation.    
In contrast, AF represents a specific and clinically relevant pathology that is commonly encountered in routine ECG diagnostic procedures. 
However, the considered datasets include a limited number of AF-positive samples; consequently, some datasets could not be evaluated for this classification task.
Finally, for SHAP feature ranking (Algorithm~\ref{alg:shap_umap}), the employed classifier C was trained on the S-sized datasets, while for visual representation purposes, when plotting reduced embedding spaces via UMAP, we selected a representative subset of 1000 balanced samples per dataset.

\subsection{Reproducibility}
To promote reproducibility, we relied on open-source datasets and open-weights FMs, except for ECG-JEPA, whose weights were provided by the authors upon request.
The framework was developed using PyTorch (v2.6.0). 
Experiments were conducted on a high-performance computing node with the following specifications: two Tesla V100-PCIE-16GB GPUs, Intel® Xeon® Gold 5118 CPU (2.30GHz), and 512GB of RAM.
We also set the random seed to 42 for all experiments.
Finally, upon possible paper publication, we plan to publicly release the entire codebase.

\newcommand{\panel}[2]{%
\begin{minipage}[t]{0.20\linewidth}
\centering
\begin{tikzpicture}
\node[inner sep=0pt] (img) {\includegraphics[width=\linewidth]{#2}};

\node[anchor=north, font=\scriptsize, yshift=-2pt]
      at (img.south) {#1};
\end{tikzpicture}
\end{minipage}
}

\begin{figure*}[t]
\centering

\includegraphics[width=0.7\textwidth]{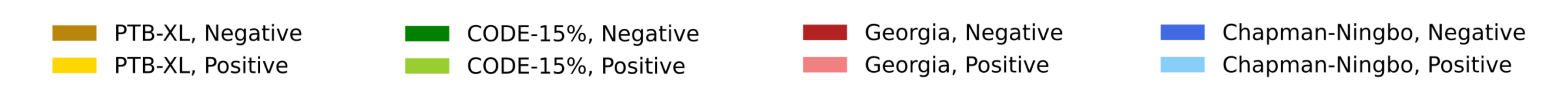}

\vspace{0.8em} 

\panel{(a) CD, ECG-FM}{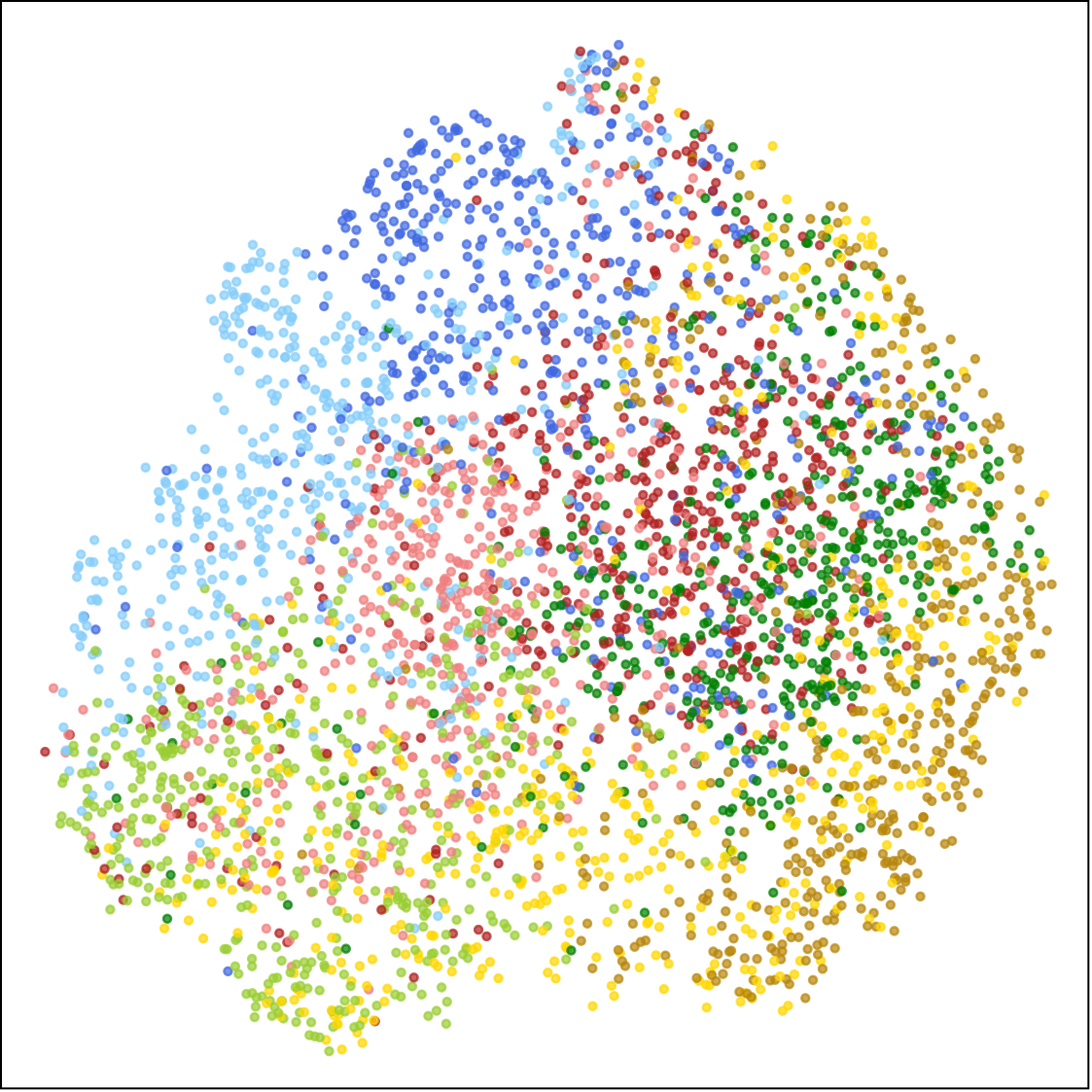}\hfill 
\panel{(d) CD, HuBERT-ECG base}{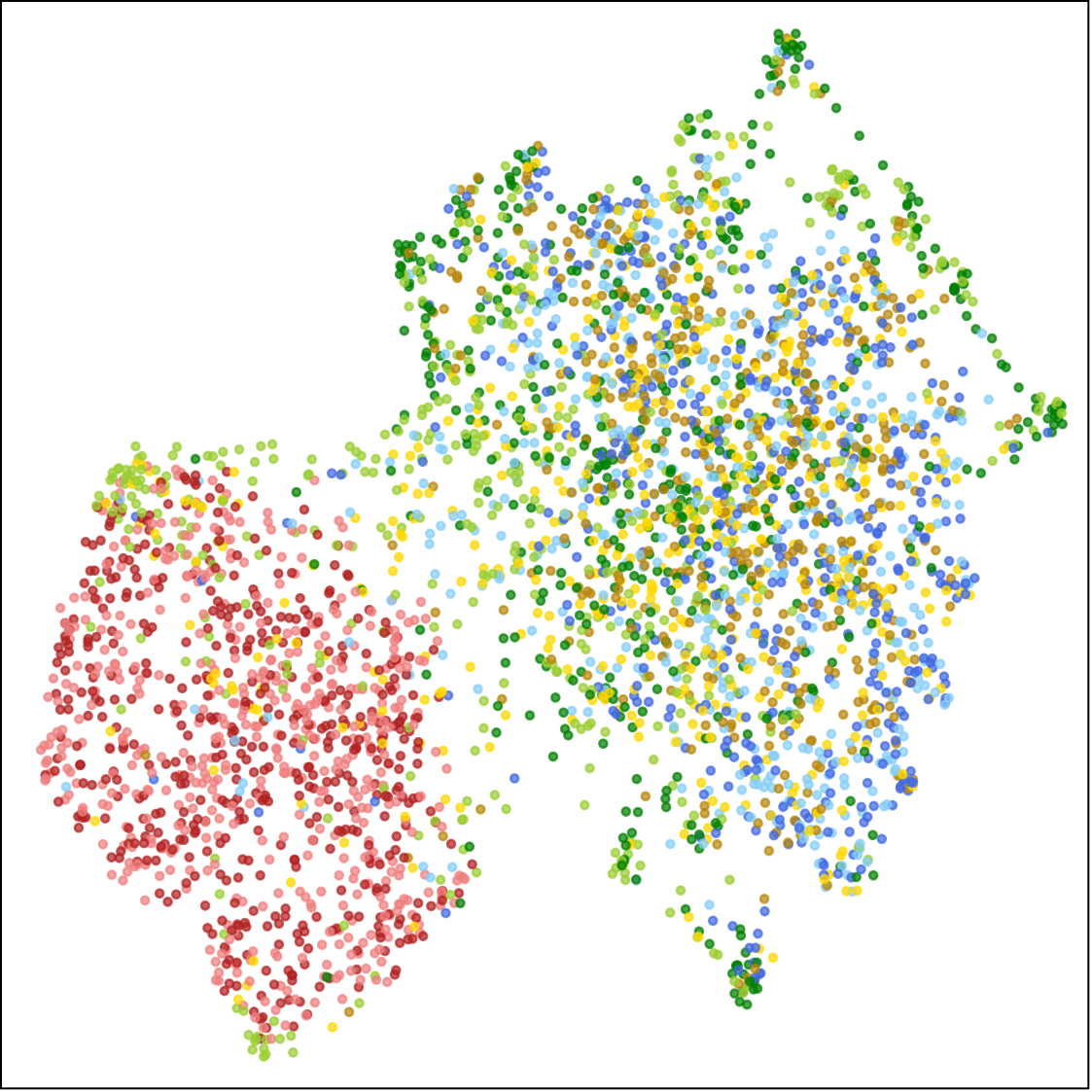}\hfill
\panel{(g) AF, ECG-FM}{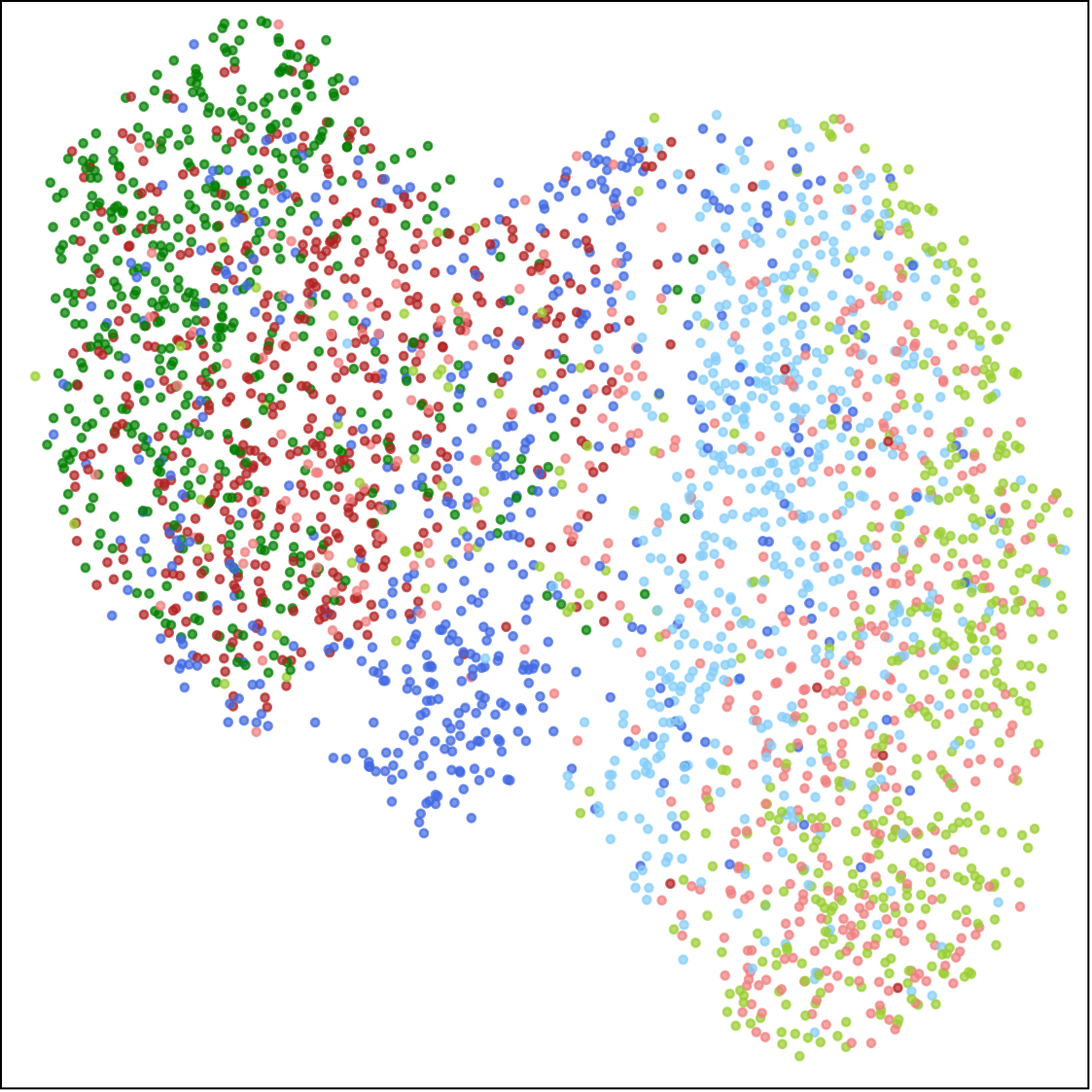}\hfill
\panel{(j) AF, HuBERT-ECG base}{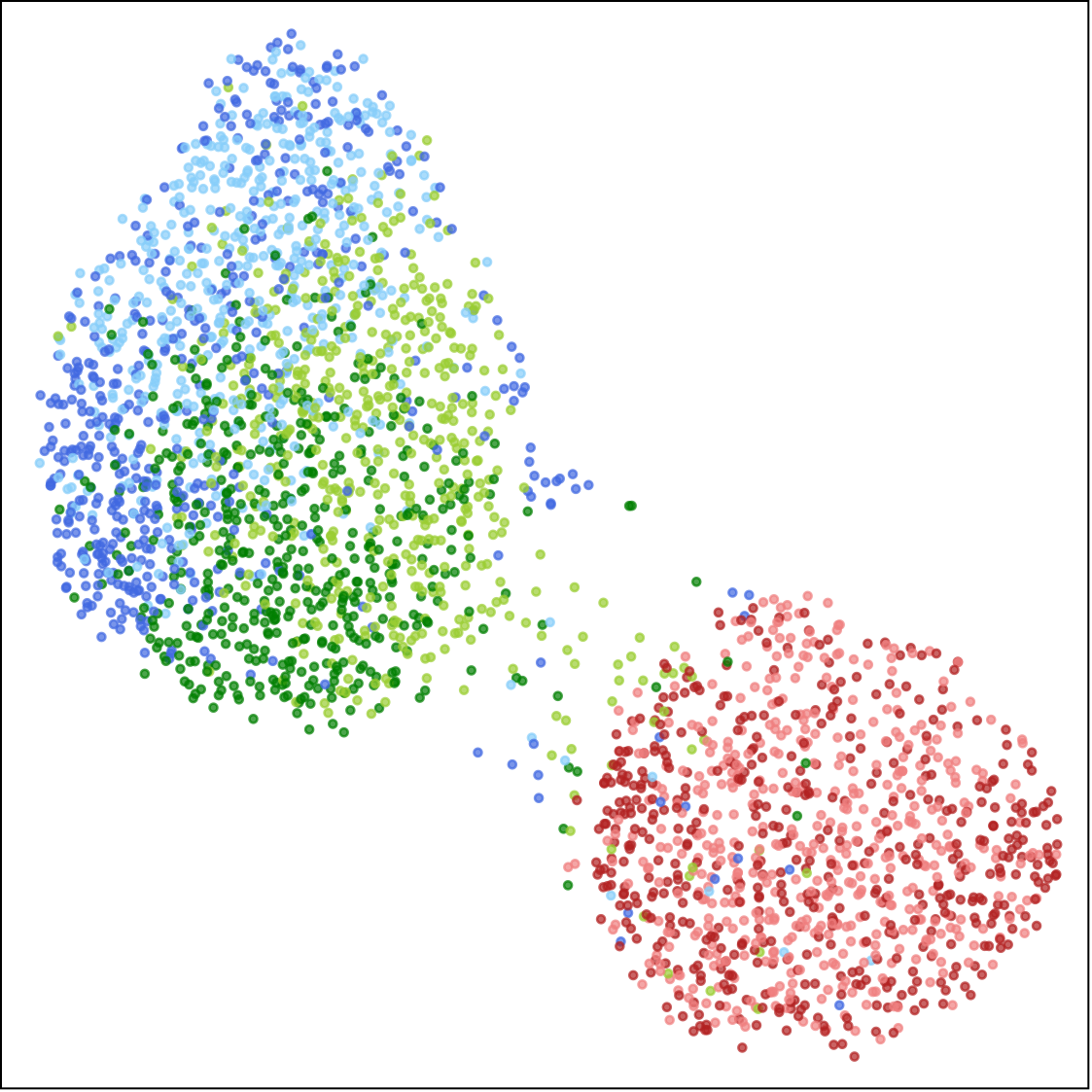}

\vspace{0.3em}

\panel{(b) CD, ECGFounder}{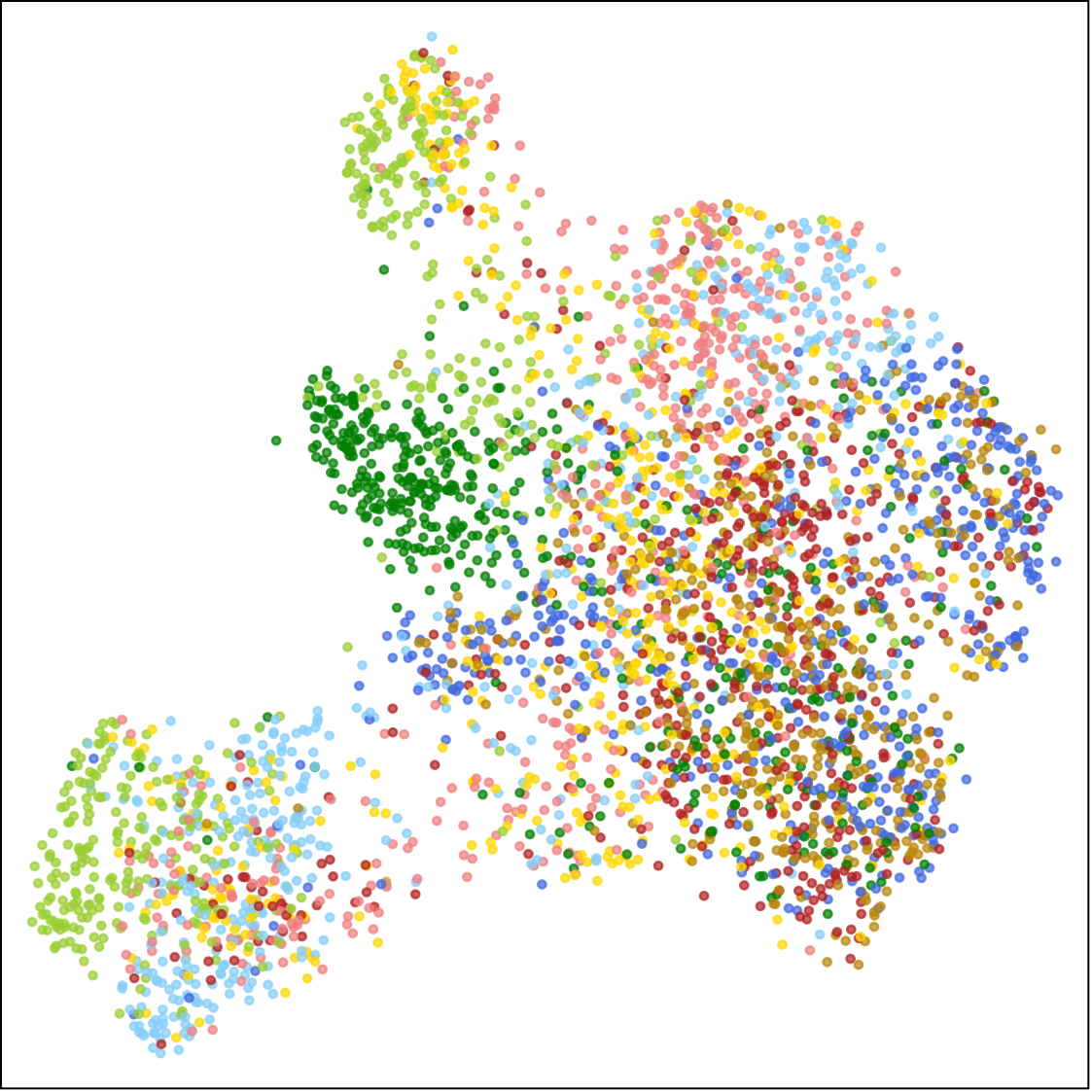}\hfill
\panel{(e) CD, HuBERT-ECG large}{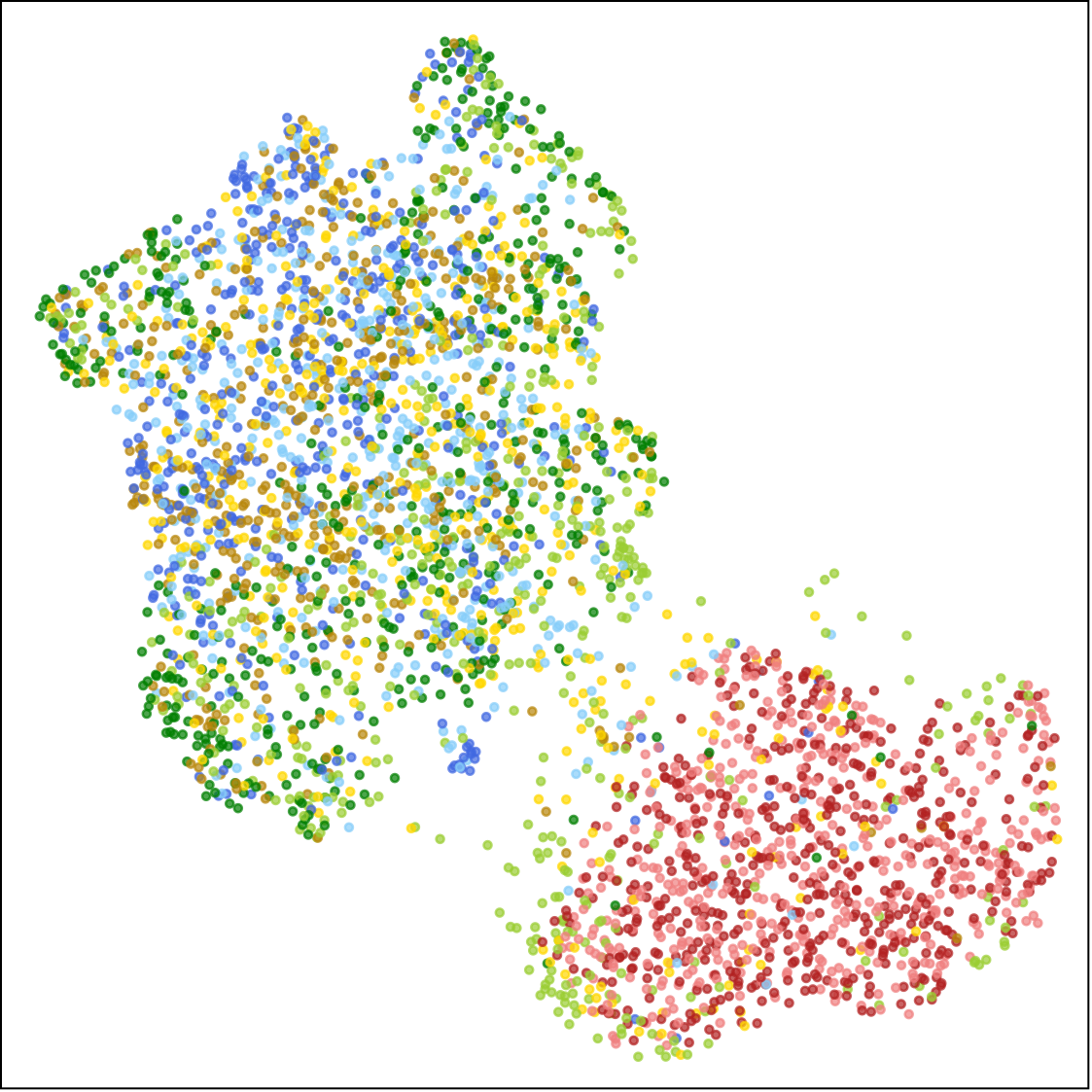}\hfill
\panel{(h) AF, ECGFounder}{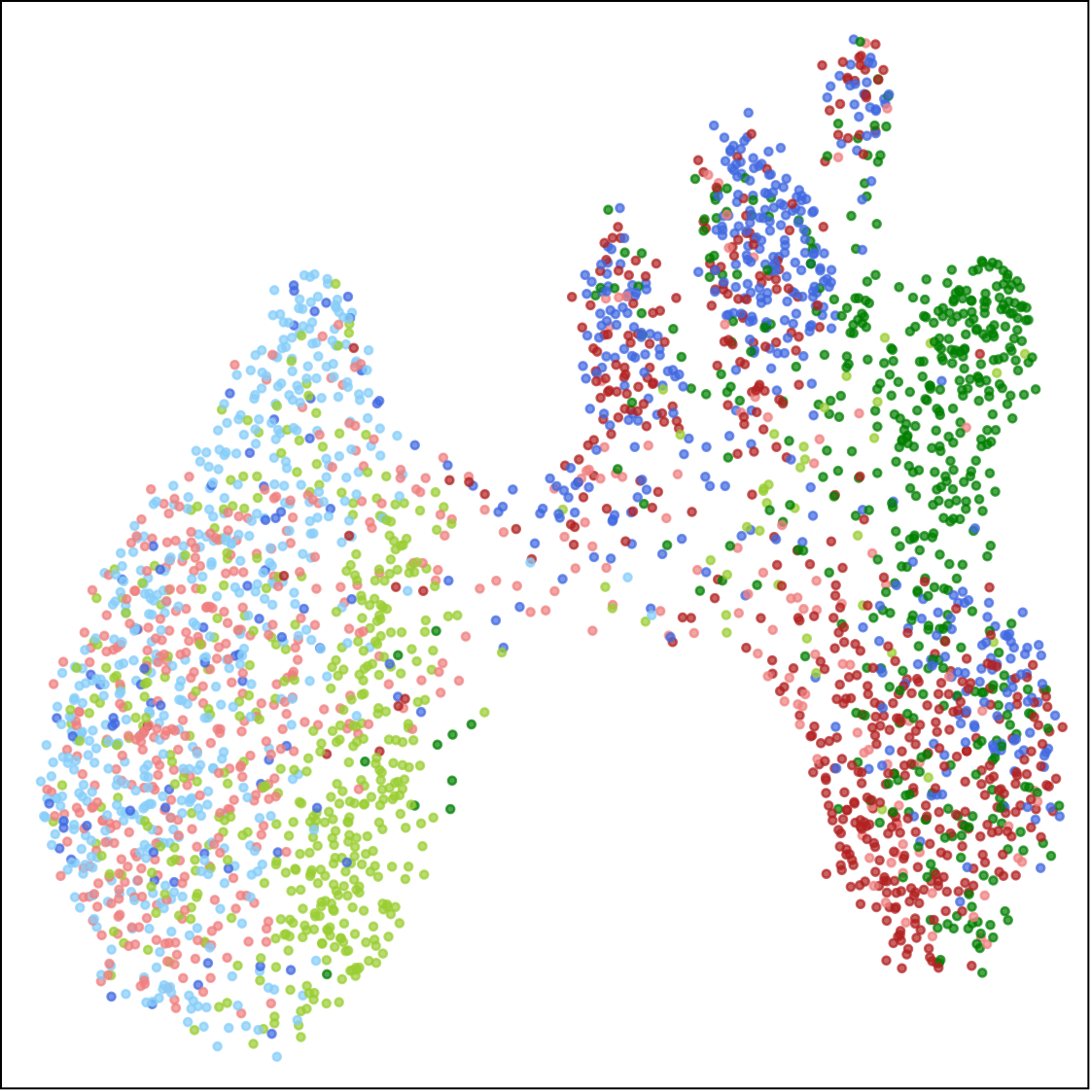}\hfill
\panel{(k) AF, HuBERT-ECG large}{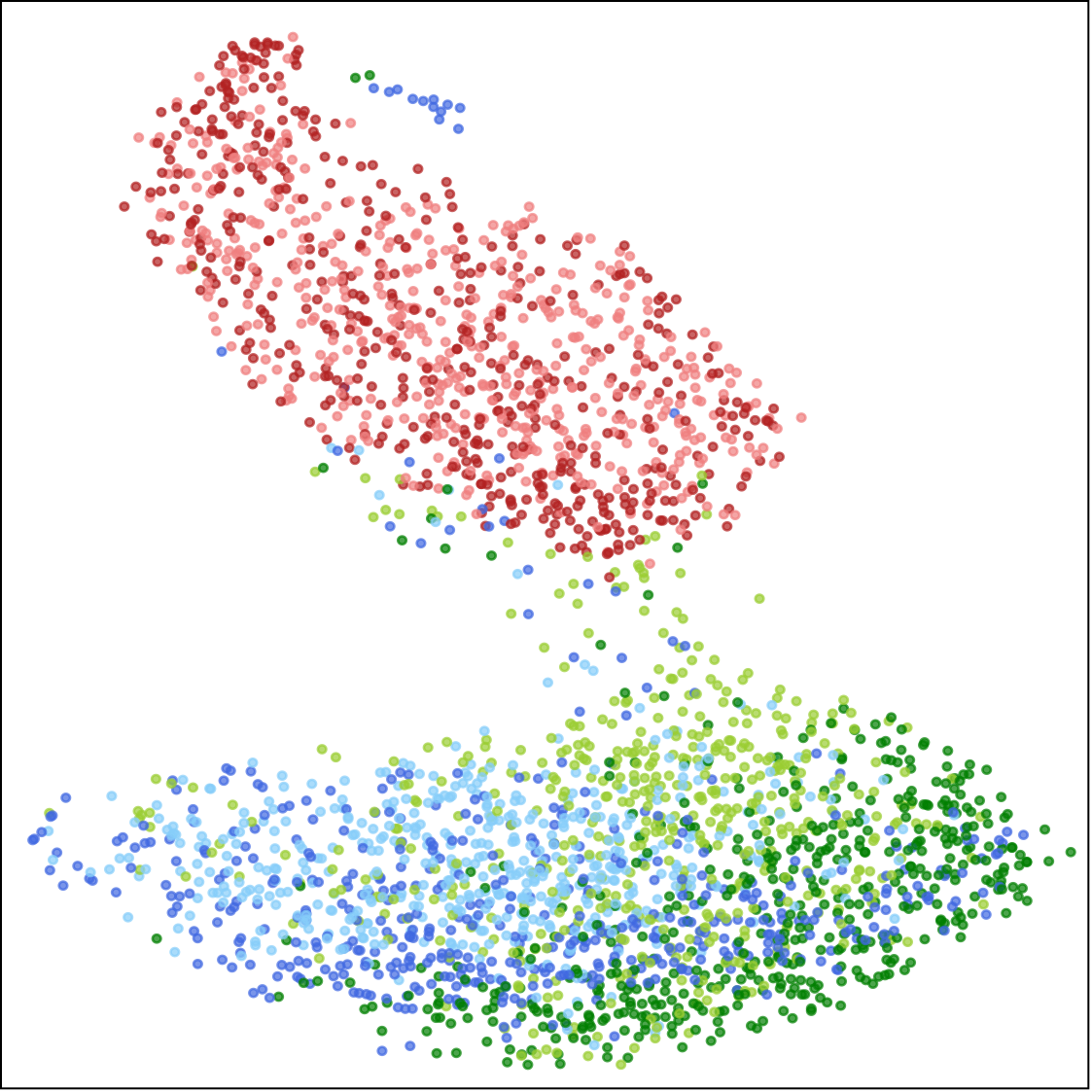}

\vspace{0.3em}

\panel{(c) CD, HuBERT-ECG small}{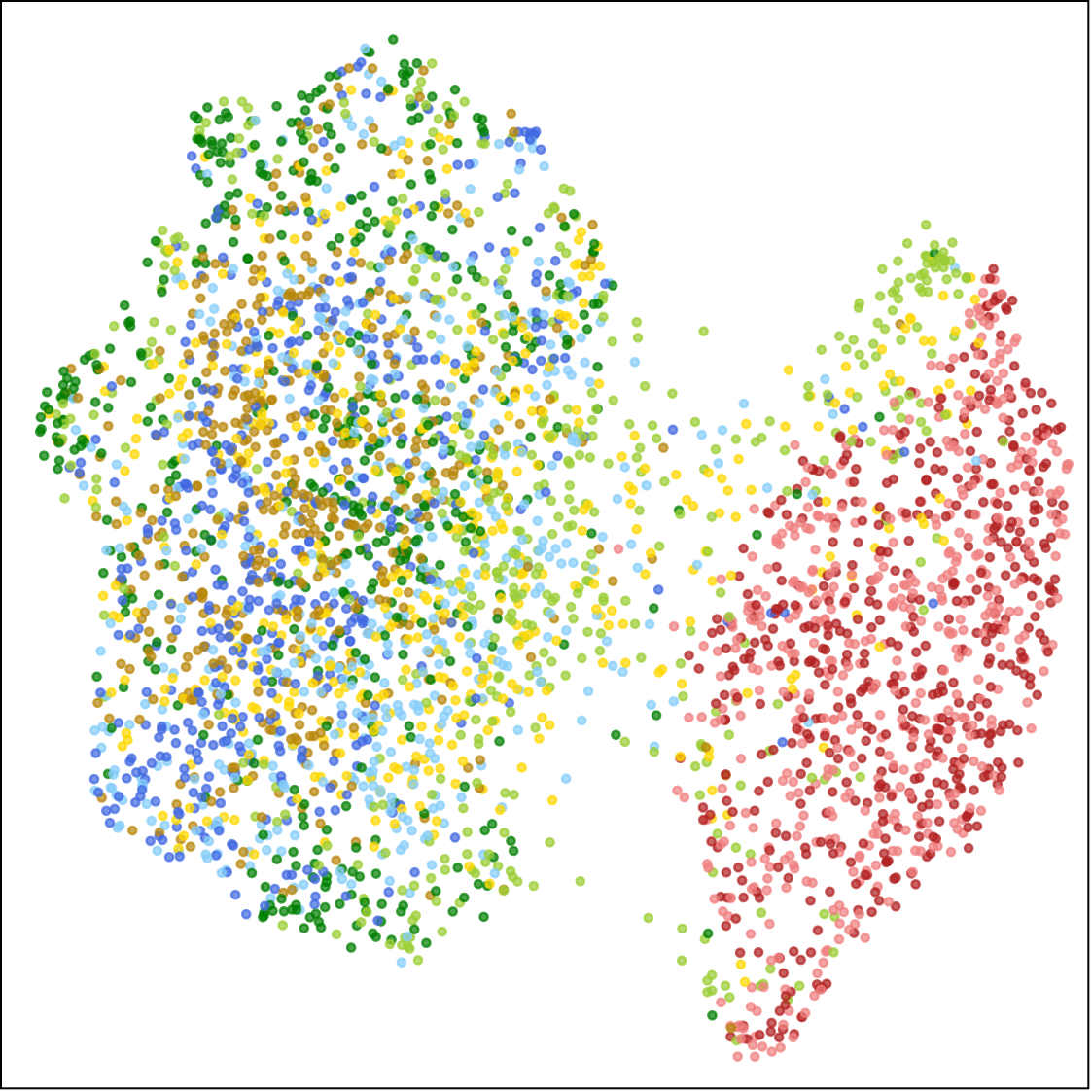}\hfill
\panel{(f) CD, ECG-JEPA}{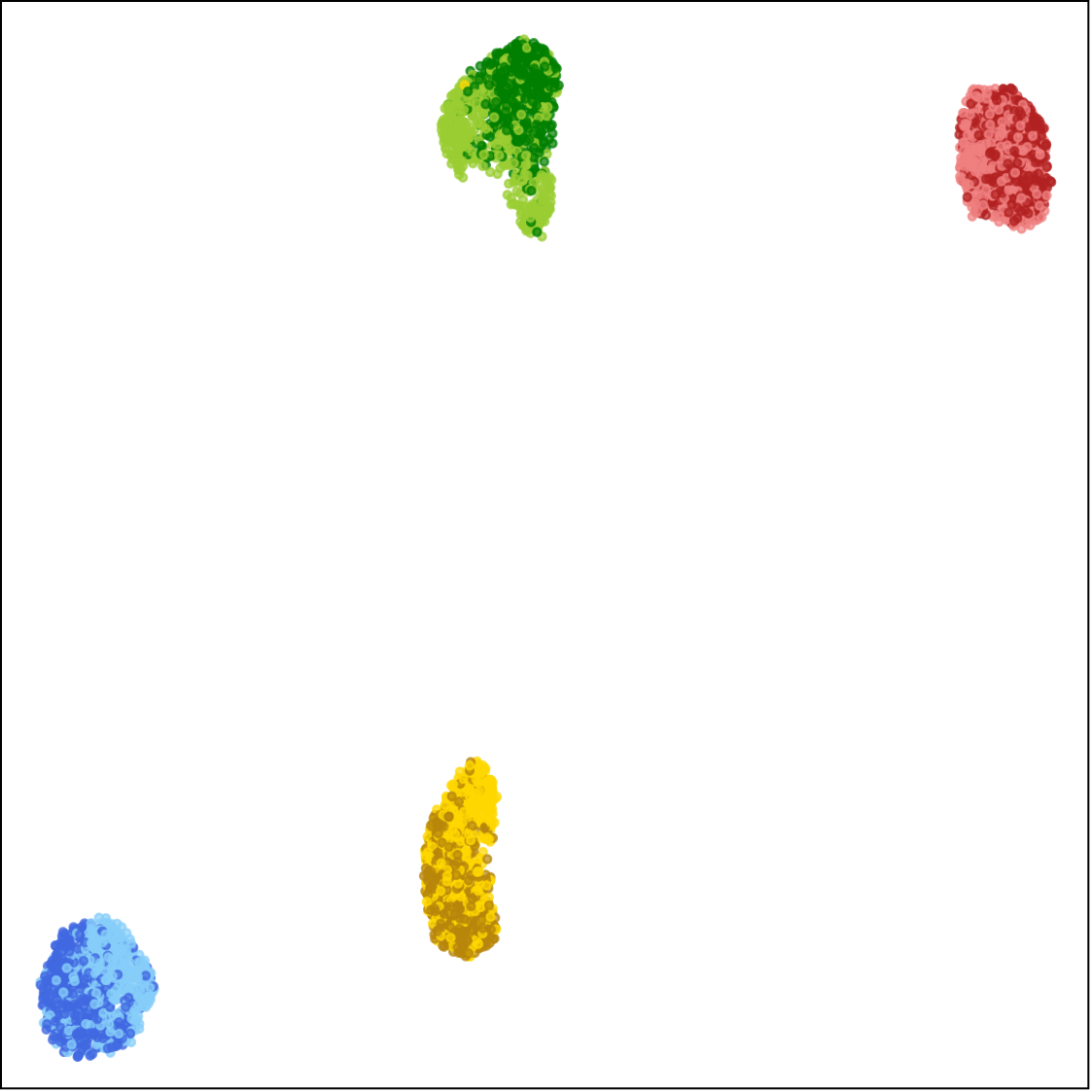}\hfill
\panel{(i) AF, HuBERT-ECG small}{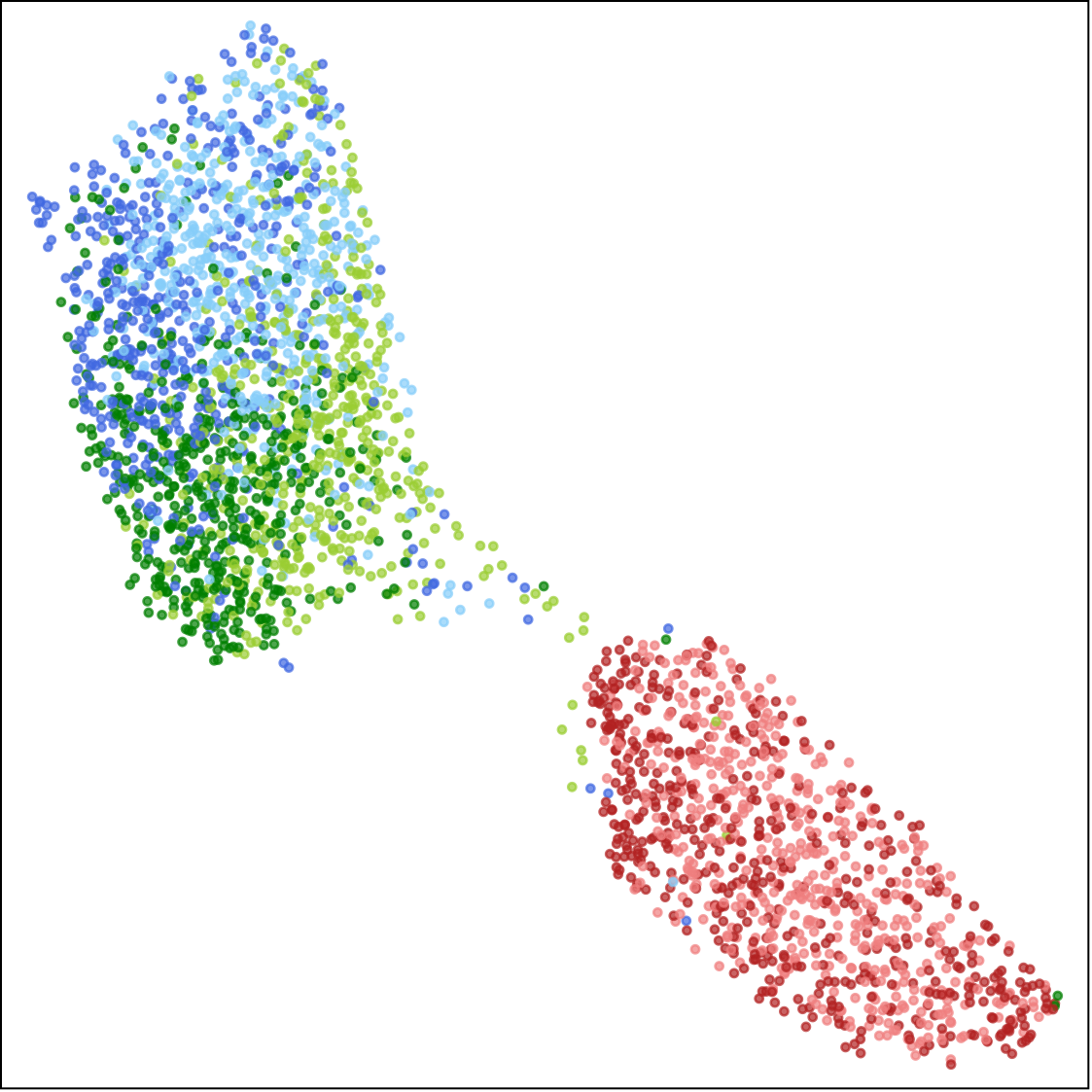}\hfill
\panel{(l) AF, ECG-JEPA}{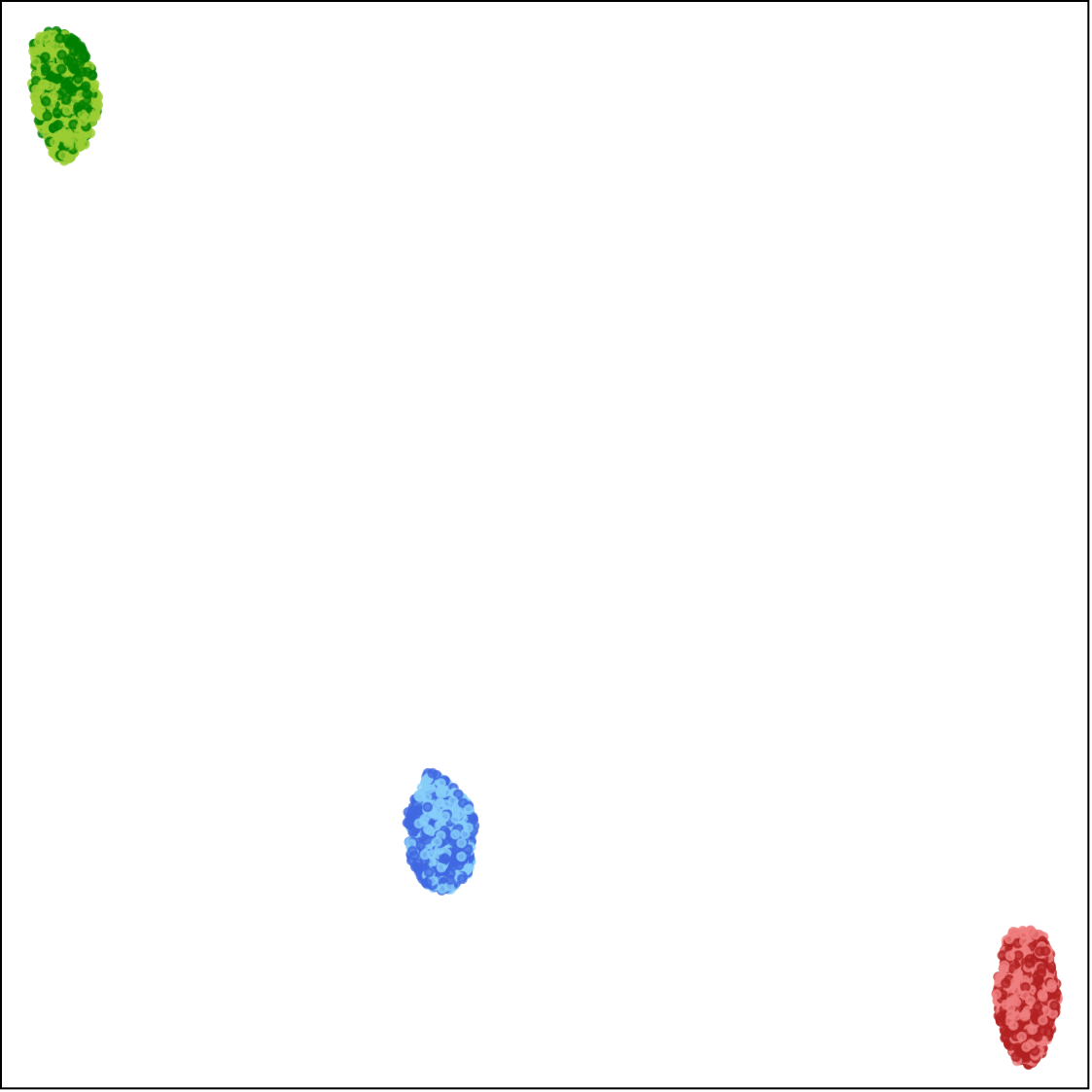}

\caption{%
UMAP-based embedding space visualization across the AF (a-f) and CD (g-l) classes for all the encompassed FMs.
Samples are colored according to  ground-truth labels.
Panels are ordered column-wise.}
\label{fig:umap_c15_cd_af}
\end{figure*}

\section{Results and Discussion}
This section presents the takeaways that emerge from benchmarking the encompassed FMs and datasets through the proposed framework.

\paragraph{Analysing Classification Performance.}
Table~\ref{tab:cd_results} reports the performance results for CD and AF classification, presenting the median F1 score. 
For each FM and label, Interquartile Ranges (IQR) are computed from the distribution of cross-validated results.
In the column header, we report the maximum IQR for each FM-class combination.
Missing values occur in those datasets not containing a sufficient number of positive instances for the target label under consideration.

First, we observe that performance over the C15 dataset is systematically higher than  other datasets regardless the FM: consequently, the $\arg\max$ operation at line~\ref{best_dataset_classifier} in Algorithm~\ref{alg:shap_umap} constantly selects C15 for each FM.
Furthermore, we notice that ECG-FM and ECGFounder consistently achieve the highest classification performance w.r.t. other FMs, even under data scarcity conditions.
These results suggest that such FMs generalize effectively regardless of dataset size, as performance rapidly saturates with increasing training data.
Nonetheless, despite the small margin, ECGFounder exhibits slightly superior performances compared to ECG-FM across all the considered settings.
On the other hand, HuBERT-ECG-based models and ECG-JEPA consistently perform worse on classification over both AF and CD.
Particularly, all HuBERT-ECG-based FMs exhibit degraded performances over the GEO dataset when compared to other datasets.
Moreover, results do not scale over HuBERT-ECG model sizes: in fact, despite increasing the model parameters, we do not observe significant increasing performance.
Finally, ECG-JEPA performs similarly to HuBERT-ECG models over the CD class, while degrading its performances over the AF class, despite it being a more clinically specific.
This behaviour is in striking contrast with the other FMs, and thus suggests the presence of some intrinsic differences between ECG-JEPA and the remaining FMs.
However, performance alone is not sufficient to understand why this happens:  therefore, a deeper investigation of the learned embeddings through representation analysis is required.

\paragraph{Analyzing Cross-Dataset Shared Top Features.}
For each FM, Figure~\ref{fig:shap_features_percentage} depicts the shared top-50 features rates among different datasets embedded by the considered FM.
The top-50 features are selected as discussed in Section~\ref{par:shap_feature_ranking}.

We observe higher shared features rates for those FMs (ECG-FM and ECGFounder) which achieve superior downstream performances, with the exception of ECGFounder over the AF label.
We hypothesize that ECGFounder achieves strong classification performance over the AF class despite the limited overlapped features, as its strong generalization capability allows even a 36.0\% share of top-ranked features to be sufficient for effective performance.

From these findings, we observe that the greater the overlap of common features across datasets, the more the FM better embeds the semantics lying behind the considered clinical class, discarding dataset-related features while retaining class-informative ones.
Consequently, this suggests relevant generalization capabilities which fairly boost lightweight downstream classifiers to achieve enhanced downstream performances despite their simple architectures, thus confirming what aforestated in Section~\ref{par:shap_feature_ranking}.

\begin{table*}[t]
\centering
\resizebox{\textwidth}{!}{
\begin{tabular}{
    l
    c>{\columncolor{afgray}}c
    c>{\columncolor{afgray}}c
    c>{\columncolor{afgray}}c
    c>{\columncolor{afgray}}c
    c>{\columncolor{afgray}}c
    c>{\columncolor{afgray}}c
    }
\toprule
\textbf{FM} 

& \multicolumn{2}{c}{\textbf{kNN@10}$_{\text{Label}} \uparrow$ }
& \multicolumn{2}{c}{\textbf{Centroid sep}$_{\text{Label}} \uparrow$ }
& \multicolumn{2}{c}{\textbf{ARI}$_{\text{Label}} \uparrow$ }
& \multicolumn{2}{c}{\textbf{kNN@10}$_{\text{Dataset}} \downarrow$}
& \multicolumn{2}{c}{\textbf{Centroid sep}$_{\text{Dataset}} \downarrow$}
& \multicolumn{2}{c}{\textbf{ARI}$_{\text{Dataset}} \downarrow$} \\
\cmidrule(lr){2-7}
\cmidrule(lr){8-13}
& \textbf{CD} & \textbf{AF}
& \textbf{CD} & \textbf{AF}
& \textbf{CD} & \textbf{AF}
& \textbf{CD} & \textbf{AF}
& \textbf{CD} & \textbf{AF}
& \textbf{CD} & \textbf{AF} \\
\midrule

ECG-FM           
& 0.75 & 0.86 & 2.43 & 4.50 & 0.23 & 0.64
& 0.80 & 0.76 & 2.78 & 2.82 & 0.31 & 0.20 \\

ECGFounder       
& \textbf{0.79} & \textbf{0.88} & \textbf{3.17} & \textbf{6.23} & \textbf{0.24} & \textbf{0.70}
& \textbf{0.71} & \textbf{0.72} & \textbf{2.34} & \textbf{2.76} & \textbf{0.03} & \textbf{0.12} \\

HuBERT-ECG small 
& 0.62 & 0.65 & 2.16 & 1.97 & 0.12 & 0.00
& 0.82 & 0.91 & 4.49 & 6.90 & 0.45 & 0.71 \\

HuBERT-ECG base  
& 0.57 & 0.63 & 1.62 & 1.84 & 0.02 & 0.04
& 0.81 & 0.90 & 4.38 & 5.33 & 0.47 & 0.69 \\

HuBERT-ECG large 
& 0.58 & 0.65 & 1.45 & 1.85 & 0.03 & 0.01
& 0.82 & 0.90 & 4.36 & 6.94 & 0.42 & 0.66 \\

ECG-JEPA         
& 0.66 & 0.57 & 2.30 & 1.30 & 0.07 & 0.00
& 1.00 & 1.00 & 7.19 & 8.45 & 0.98 & 1.00 \\

\bottomrule
\end{tabular}
}
\caption{Label-level and dataset-level separability across FMs' embedding spaces.
Metrics are reported separately for CD and AF classes.
For optimal embedding generalizablity, metrics marked with ($\uparrow$) should be maximized, whereas those marked with ($\downarrow$) should be minimized.
In bold we mark the optimal score across FMs for each class.}
\label{tab:latent_separability}
\end{table*}

\paragraph{Analyzing Embedded Representations.}
Across all models, ECGFounder and ECG-FM correctly optimize the dataset-level and label-level separability (Table~\ref{tab:latent_separability}), tending to minimize the first while maximizing the second.
Despite disposing over different geometries, such patters are identifiable also from the UMAP-based embedding space visualization: in fact, different classes are mapped to different areas of the embedding space.
This distinction is particularly evident especially over the AF class (Figure~\ref{fig:umap_c15_cd_af}.g-~\ref{fig:umap_c15_cd_af}.l).
Overall, this indicates more dataset-invariant and clinically aligned representations.

Conversely, HuBERT-ECG and ECG-JEPA exhibit a systematically higher dataset-level separability w.r.t. label-level separability, particularly when considering kNN agreement and ARI (Table~\ref{tab:latent_separability}).
This confirms that such model retain dataset-dependent information, thus suggesting that the global organization of the embedding space is driven by dataset-specific characteristics rather than by the clinical label.
Moreover, HuBERT-ECG models display a clear size-dependent trend: as model size increases, dataset-level separability becomes dominant while label-level alignment degrades.
UMAP representations confirm this by placing samples from GEO in a distinct cluster (Figures~\ref{fig:umap_c15_cd_af}.c-~\ref{fig:umap_c15_cd_af}.e, ~\ref{fig:umap_c15_cd_af}.i-~\ref{fig:umap_c15_cd_af}.k), thus increasing dataset-level separability, while contextually spreading labels across the whole embedding space.
Finally, ECG-JEPA constitutes an extreme case, achieving near-perfect dataset-level separability while providing limited label discrimination. 
This mirrors what is visualized through UMAP, where dataset-based clusters are clearly identifiable (Figures~\ref{fig:umap_c15_cd_af}.f, \ref{fig:umap_c15_cd_af}.l).

Importantly, these findings coherently align with the observer downstream performances: higher label-level separability in the embedding space is consistently associated with classifiers' improved performance, suggesting that clinically aligned representations provide more discriminative and informative features.

\paragraph{Examining Embeddings’ Informativeness.}
From the performance and representation evaluation we conducted in this benchmark, it emerges that ECG-FM and ECGFounder outperform other FMs.
On one hand, this highlights that architectures based on CNN backbones, such as RegNet (ECGFounder), are particularly effective at capturing the local morphological patterns that characterize ECG signals. 
On the other hand, it underscores that hybrid CNN–Transformer models (ECG-FM, HuBERT-ECG) can further enhance representation learning by integrating long-range temporal dependencies; however, their effectiveness critically depends on the adopted pretraining strategy.
In ECG-FM, continuous objectives allow to preserve fine-grained signal information while leveraging global contextual modeling; in contrast, HuBERT-ECG explicitly discretizes the embedding space, which may limit the retention of subtle but important details.
Moreover, this empirically proves that it is not always true that large-scale or highly parametrized FMs lead to more informative embeddings.

Importantly, among all the datasets, PTX emerges as a particularly challenging one for FMs to embed, as evidenced by both degraded performance and embedded representation evaluation. 
We hypothesize this stems from the broader (and more complex) set of CD subclasses this dataset includes (see Table~\ref{tab:cd_subclasses}).
Nonetheless, the PTX-based observations enable us to conclude that even FMs performing well on some classes tend to exhibit a degradation in embedding informativeness when dealing with less recurrent pathological conditions.

\section{Conclusion}
In this work, we propose a structured methodology to benchmark ECG-expert FMs.
One of the main advantages of our benchmarking framework, we mention the fact that it complements embedded representation analysis to the more traditional performance analysis, thus filling a significative gap in the current ECG-expert FM benchmarks literature.
Moreover, it is cross-continental and State-Of-The-Art-aligned, as it evaluates ECG-expert FMs pretrained with up-to-date techniques over a dataset cohort composed of different continent-sourced ECG signals.
Furthermore, we also include real-world scenario evaluation, as the encompassed FMs were tested over different sized dataset subsets, considering also a data scarcity setting, which is a rather common real-world scenario in the healthcare domain.
The present study confirms that performance analysis alone is not enough to evaluate FMs' generalization capabilities.
In fact, while performance may suggest preliminary hints, true confirmation only comes from understanding the embedded representations' structure: performance shows \textit{what} embeddings achieve, whereas representation analysis reveals \textit{how} they are built.
Specifically, from our experiments we observe that those embeddings which preserve dataset-related structure tend to exhibit limited label separability (HuBERT-ECG, ECG-JEPA), whereas dataset-independent embeddings lead to improved clinical alignment (ECGFounder, ECG-FM).
This experimental evidence further confirms that a benchmarking methodology like the one herein proposed is essential for effective and in-depth understanding of ECG-expert FMs' embeddings informativeness.

Furthermore, the work can be enriched and improved.
Indeed, some of the evaluated FMs were pretrained on part of the datasets employed in this benchmark.
However, avoiding this overlap was not feasible as, to the best of our knowledge, no open-source out-of-distribution ECG dataset is currently available w.r.t. the encompassed FMs. 
As such, future works may focus on addressing this issue.
Moreover, we may also investigate the behaviour of FMs across a broader range of cardiac pathologies, in order to assess whether the observed trends generalize to other pathological conditions.

\section*{Ethical Statement}
We benchmark frozen ECG foundation models using only publicly available research datasets under their licenses (privacy/copyright/consent), collect no new data, and attempt no re-identification. Results are for research benchmarking only (not direct clinical use); any deployment would require prospective validation, IRB/regulatory oversight, and safety/fairness assessment. We will release code/configs, but not redistribute datasets or non-redistributable weights.

\bibliographystyle{named}
\bibliography{ijcai26}

\end{document}